\documentclass[a4paper]{llncs} 
\pdfoutput=1

\usepackage[english]{babel}
\usepackage{lmodern}
\usepackage[T1]{fontenc}
\usepackage[utf8]{inputenc}
\usepackage[pdftex]{graphicx}
\usepackage{hyperref}
\usepackage[export]{adjustbox}
\usepackage[babel=true]{microtype}
\usepackage{tabularx}
\usepackage{ragged2e}
\usepackage[backend=bibtex,style=lncs]{biblatex}
\usepackage{booktabs}
\usepackage{csquotes}

\usepackage[style=base, labelfont=bf, labelsep=period, tableposition=below]{caption}
\usepackage[labelfont=bf]{subcaption}

\setkeys{Gin}{width=\linewidth}

\graphicspath{{img/}}

\usepackage[%
rm={oldstyle=false,proportional=true},%
sf={oldstyle=false,proportional=true},%
tt={oldstyle=false,proportional=true,variable=true},%
qt=false%
]{cfr-lm}

\usepackage{xspace}

\newcommand{\ie}{i.\,e.,}

\makeatletter
\g@addto@macro{\UrlBreaks}{\UrlOrds}
\makeatother

\usepackage[capitalise]{cleveref}
\crefname{section}{Sect.}{Sect.}
\Crefname{section}{Section}{Sections}
\crefname{subsection}{Subsect.}{Subsect.}
\Crefname{subsection}{Subsection}{Subsections}

\title{Is This a Joke? Detecting Humor in Spanish Tweets}

\author{Santiago Castro \and Matías Cubero \and Diego Garat \and Guillermo Moncecchi}

\institute{
	Universidad de la República, \\
	Montevideo, Uruguay \\
	\email{\{sacastro, mcubero, dgarat, gmonce\}@fing.edu.uy}
}

\begin{document}

\maketitle

\begin{abstract}
	While humor has been historically studied from a psychological, cognitive and linguistic standpoint, its study from a computational perspective is an area yet to be explored in Computational Linguistics. There exist some previous works, but a characterization of humor that allows its automatic recognition and generation is far from being specified. In this work we build a crowdsourced corpus of labeled tweets, annotated according to its humor value, letting the annotators subjectively decide which are humorous. A humor classifier for Spanish tweets is assembled based on supervised learning, reaching a precision of 84\% and a recall of 69\%.

\end{abstract}

\section{Introduction}

The human being as a species is characterized by laughter. Humor, which is a potential cause of laughter, is an essential component of human communication. Not only does it allow people to feel comfortable, but also produces a cozier environment. While humor has been studied from a psychological, cognitive~\cite{humorJournal} and even linguistic~\cite{raskin1985semantic} standpoint, its study from a computational viewpoint is still an area to be explored within Computational Linguistics. There exist some previous works~\cite{so63066}; however, a humor characterization that allows its automatic recognition and generation is far from being specified, particularly for the Spanish language.

Identifying humor in a text can be seen as an intermediate step for the resolution of more complex tasks. It would be interesting to generate jokes, or humor in general, based on the knowledge of which attributes enrich texts in a better way. Another appealing use case is to exploit the outcome of a humor detector to decide automatically if a text span can be taken seriously or not. On the other hand, by way of a more direct use, humor identification can be used to find jokes on Twitter, to search for potentially funny tweets about certain trending topic or to search for humorous answers to comments on the social network.

We address herein the problem of detecting humor in Spanish tweets. It should be noted that this is different from trying to recognize humor in arbitrary texts, due to tweets' length. Here it could be assumed that tweets are either humorous or not, but not both, because they are brief (up to 140 characters). This is not always the case in others texts, as jokes could only exist in some parts but not on the whole text. Another advantage considered is that there are plenty of tweets available to analyze.

Since there is no clear definition of what humor is, how can we detect something that is in principle vaguely stated? We explore different ideas, and we finally decide to let people define it themselves by voting tweets from a web page and an Android app, in which they can label a tweet as humorous or not humorous. Once we have defined which tweets are humorous, we tackle the problem of humor detection using a supervised learning approach. In other words, we infer a function that identifies humor from labeled data. We use several techniques such as Support Vector Machine, Nearest Neighbors, Decision Trees and Naive Bayes. In order to build a set of features, we first study the state of the art of the Computational Humor area, focused on recognition and in Spanish.

In \cref{sec:computational-humor} we present the humor detection problem and its state of the art, including features studied in previous works. In \cref{subsec:corpus} we show the corpus built for this purpose and in \cref{subsec:classifier} we describe the classifier used. Afterwards, we present an experimental evaluation in \cref{sec:evaluation} and finally the conclusions in \cref{sec:conclusions}.

\section{Computational Humor}
\label{sec:computational-humor}

Computational Humor is a recent field of study about recognizing and generating humor through automatic processing. The task of language understanding is rather hard, and so are tasks related to humor. Furthermore, humor entails the usage of figurative language, which obviously makes language handling harder.

Humor by itself is not a clearly determined concept. According to Real Academia Española\footnote{\url{http://dle.rae.es/}}, humor is defined as a way of presenting reality, highlighting the comic or ridiculous side. As for comedy, it is a kind of drama meant to cause laughter. However, what causes laughter? There are several theories which try to answer this question, and consequently attempt to find what humor is. A report on the state of the art about Humor and Computational Humor~\cite{so63066} enumerates some of them. The main ideas of these theories are described hereinafter. Readers will notice that these ideas are similar, in spite of putting the focus on different attributes.

\textcite{gruner2000game} develops a theory which claims that humor is related to superiority feelings, asserting that there is always a winner in every joke. \textcite{freud, Minsky80jokesand} state that humor is about relieving repressed feelings. In this case, laughter relieves the stress caused by taboo topics, such as death, marriage or sex. The Theory of the Incongruity Resolution~\cite{rutter1997stand} claims that two objects are presented under the same concept, with details applying to both and with similarities, but as narration progresses it turns out that only one is possible. Furthermore, we have The Semantic Script Theory of Humor and The General Theory of Verbal Humor~\cite{attardo1991script},~\cite{ ruch1993toward}. They state that humor is about two scripts which come into conflict with each other, where there are two opposed subjects contrasted, such as big vs small, death vs life, normal vs abnormal, among others.

Let us introduce an example\footnote{Taken from \url{https://twitter.com/chistetipico/status/430549009812291584}. It has been slightly adapted to maintain an appropriate language.}:

\begin{displayquote}
  \centering
  --- Nada es imposible.

  --- A ver, tocate la espalda con la rodilla, mente positivista.

  \begin{center}
    \scriptsize
    --- Nothing is impossible.

    --- Seriously? Touch your back with your knee, you positivist mind.
  \end{center}
\end{displayquote}

Following the Superiority Theory, the reader is the winner when he laughs at the positive person, feeling superior as the latter lose the dispute. According to the Relief Theory, we laugh with the purpose of releasing tension, which in this case can be provoked by talking about the limits of life, such as when saying ``nothing is impossible''. The Theory of the Incongruity Resolution also applies here due to the fact that there is ambiguity; with ``nothing is impossible'' the example implies that all your dreams may come true, but the person is answered as if the statement was literal.

\subsection{Humor Detection}

The concrete goal of this research is to classify tweets written in Spanish as humorous or not humorous. In order to accomplish this, jokes need to be completely expressed within the text, and no further information must be required (apart from contextual information). Since Twitter allows only brief publications --- no more than 140 characters --- we freely assume the text to be a unit: either the whole tweet is humorous, or it is not.

\subsection{State of the Art}
\label{sec:state-of-the-art}

We did not find any attempt to automatically recognize humor for Spanish. Notwithstanding, \textcite{Mihalcea:2005:MCL:1220575.1220642, so63066} built humor detectors for English making use of \emph{one-liners}, \ie{} texts of approximately fifteen words. Supervised learning was used to produce an outcome --- humorous or not humorous content --- based on features which might reflect certain properties that humor should satisfy. Furthermore, \textcite{DBLP:conf/tsd/ReyesBR09, DBLP:journals/pdln/ReyesRMT09} have gathered and studied features specific to humor, without having the objective of creating a recognizer.

A concise compilation of the features presented in these works is shown below:

\begin{description}
  \item[Adult Slang:] According to~\textcite{Mihalcea:2005:MCL:1220575.1220642}, adult slang is popular in jokes. Let us remember that the Relief Theory states that laughter releases stress caused by taboo subjects, and adult slang could be one. WordNet Domains~\cite{strapparava2004wordnet} can be used to search for words tagged with the domain ``Sexuality'' in potentially humorous texts.

  \item[Alliteration:] This is about the repetition of phonemes in a text. It is a generalization of the rhyme. As stated in~\cite{Mihalcea:2005:MCL:1220575.1220642}, structural and phonetic properties of jokes are at least as important as their content.

  \item[Ambiguity:] It may be explained by the Incongruity Resolution Theory that ambiguity plays an important role, as it gives more than one interpretation to texts. \textcite{conf/wilf/SjoberghA07, Basili02parsingengineering, DBLP:conf/tsd/ReyesBR09} mention different ways to measure it, such as counting the number of meanings of the words that appear or counting the number of possible syntax trees.

  \item[Antonymy:] Following the Semantic Script Theory of Humor, we could look for opposed terms in texts, and that is how this feature is supported. The idea is to take into account pairs of antonym words mentioned in texts. Wordnet~\cite{wordnet} is useful since it is a lexical database which contains antonyms for English words, among other relations.

  \item[Keywords:] There are certain words that are more used in humorous contexts than in normal situations~\cite{conf/wilf/SjoberghA07}. An example of these are words related to animal contexts, lawyers, etc.

  \item[Language model perplexity:] In~\textcite{DBLP:conf/tsd/ReyesBR09} a language model is built from narrative texts, and perplexity\footnote{Perplexity is a measurement of how well a probability model predicts a sample. Low perplexity indicates the probability model is good at predicting the sample. It is defined as $2^{- \frac{1}{n} \sum_{i=1}^n \log_2 p(x_i)}$, where $x_1, \ldots, x_n$ are the sample data and $p(x_i)$ is the probability assigned to each one.} is used as a feature. Humorous texts have a higher perplexity than those which are not humorous.

  \item[Negativity:] There is a certain kind of humor which tends to have negative connotations~\cite{Mihalcea_characterizinghumour:}~\cite{DBLP:journals/pdln/ReyesRMT09}. It can be about denying, such as when saying ``no'', ``don't'' or ``never'', when talking about subjects with negative polarity such as ``bad'', ``illegal'' or ``wrong'' or when it is related to words referring to stressful subjects, such as ``alcohol'' or ``lie''.

  \item[People-centered words:] Humorous texts are constantly referring to scenarios related to people, with dialogues and references such as ``you'', ``I'', ``woman'' and ``my''. This is supported by~\textcite{Mihalcea_characterizinghumour:, journals/ci/MihalceaS06}.
\end{description}

\textcite{Mihalcea:2005:MCL:1220575.1220642} used the features Adult Slang, Alliteration and Antonymy, while \textcite{conf/wilf/SjoberghA07} focused on Alliteration, Ambiguity, Keywords and People-centered words. Both studies collected humorous one-liners from the Internet. \textcite{conf/wilf/SjoberghA07} employed only the British National Corpus (BNC) as negative samples whereas \textcite{Mihalcea:2005:MCL:1220575.1220642} additionally used proverbs and news headlines from Reuters. In both works they tried with Naïve Bayes and Support Vector Machine classifiers, resulting in no significant difference between these techniques. On one hand, \textcite{Mihalcea:2005:MCL:1220575.1220642} achieved their best accuracy with headlines: 96.85\%, while they reached 84.82\% with proverbs and 79.15\% with the BNC\@. Alliteration proved to be the most accurate feature. On the other hand, \textcite{conf/wilf/SjoberghA07} achieved an accuracy of 85.40\%, with Keywords being the most useful. \Cref{tab:comparison} summarizes the main differences and compares both studies.

\begin{table}
  \caption{Comparison of the approach of both works. The results are not directly comparable as they use different corpora.}
  \begin{tabular}{>{\Centering}m{1.8cm} >{\Centering}m{5.0cm} >{\Centering}m{5.0cm}}
    \toprule
    & \multicolumn{1}{c}{\textcite{Mihalcea:2005:MCL:1220575.1220642}} & \multicolumn{1}{c}{\textcite{conf/wilf/SjoberghA07}} \\
    \midrule
    Negative samples & BNC sentences, news headlines and proverbs & Other sentences from BNC \\
    \midrule
    Accuracy & 96.95\% with headlines, 79.15\% with the BNC and 84.82\% with the proverbs & 85.40\% \\
    \midrule
    Features & Adult Slang, Alliteration and Antonymy & Alliteration, Ambiguity, Keywords and People-centered words \\
    \bottomrule
  \end{tabular}
\label{tab:comparison}
\end{table}

\section{Proposal}

\subsection{Corpus}
\label{subsec:corpus}

Our first goal is to build a corpus with samples of humorous and non-humorous tweets. Based on~\textcite{Mihalcea:2005:MCL:1220575.1220642}, we choose to use non-humorous sample tweets that fall into the following topics: news, reflections and curious facts. For humorous samples, we extracted tweets from accounts which appeared after having searched for the keyword ``chistes'' (``jokes'' in Spanish). In total, 16,488 tweets were extracted from humorous accounts and 22,875 from non-humorous. The two groups are composed of 9 Twitter accounts each, with the non-humorous containing 3 of each topic. The amount of tweets in each topic is similar.

We tagged all tweets from news, reflections and curious facts as non-humorous, as random sampling showed that there was no humor in them. Conversely, not all tweets that were extracted from a humorous account were in fact humorous. Many of them were used to increase their number of followers, to express their opinion about a fact or to support a cause through retweets.

\begin{figure}
  \includegraphics[frame]{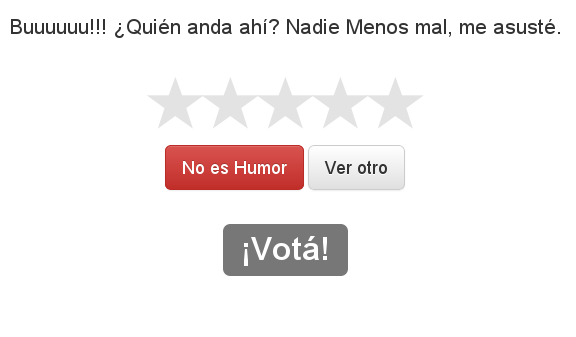}
  \caption{Page used to annotate tweets, with an example tweet on screen.}
\label{fig:page}
\end{figure}

A crowdsourced web\footnote{\url{http://clasificahumor.com}} and a mobile\footnote{\url{https://play.google.com/store/apps/details?id=com.clasificahumor.android}} annotation was carried out in order to tag all tweets from humorous accounts. In order to obtain as many annotations as possible, we wanted to keep it simple. Therefore, we showed random tweets to annotators (avoiding duplicates), providing no instructions, and let them implicitly define what humor is. In addition, the user interface was simple, as shown in \cref{fig:page}. The users could either provide a ranking of humor between one and five, express that the tweet was not humorous or skip it.

\begin{figure}
  \includegraphics{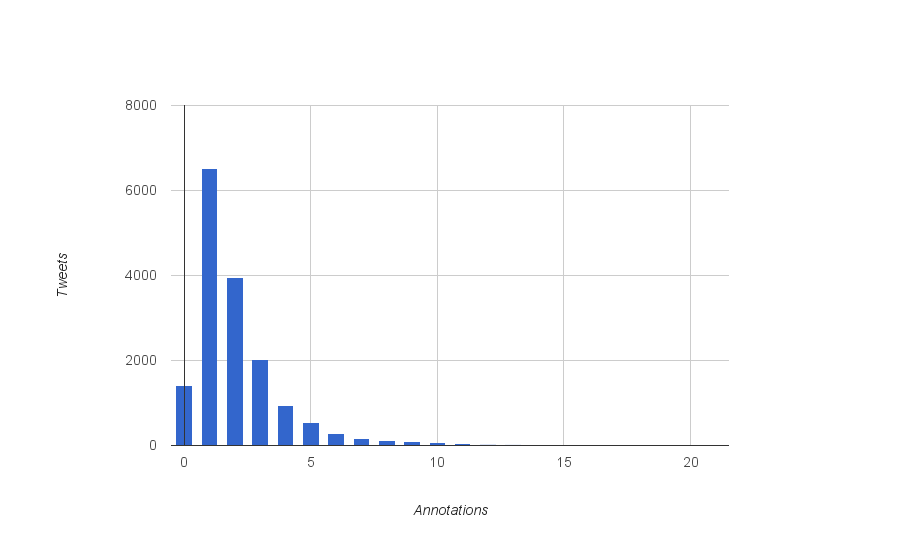}
  \caption{Histogram of annotations. Note that most tweets have few annotations.}
\label{fig:histogram}
\end{figure}

In total, 33,531 annotations were achieved, after filtering some of them that occurred in a short time lapse in the same session and with the same tag. About half of the labels were non-humorous, while the other half was divided approximately between the five rankings. A histogram of the annotations is shown in \cref{fig:histogram}. Regarding the agreement among annotators, the Fleiss' Kappa measurement for tweets with 2 annotations\footnote{Note that Kappa assumes a fixed number of annotators. For this reason, we measure it with 2 and 6, in order to give an idea of the agreement having a value with many tweets but few annotators, and other value with few tweets but many annotators.} is 0.416 and for those with 6 annotations it is 0.325.

\begin{figure}
  \centering

  \begin{subfigure}[t]{.45\textwidth}
    \centering
    \includegraphics[width=\textwidth]{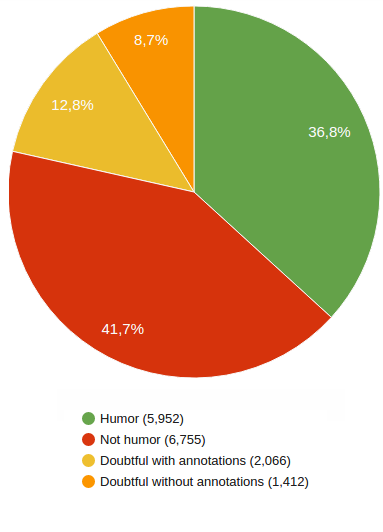}
    \caption{Graph showing the percentage of tweets from humorous accounts in each category.}
\label{subfig:decision}
  \end{subfigure}\hfill
  \begin{subfigure}[t]{.45\textwidth}
    \centering
    \includegraphics[width=\textwidth]{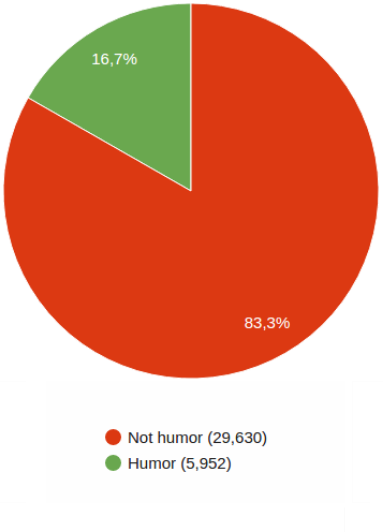}
    \caption{Pie displaying the ratio between positives and negatives in the corpus, after the decision was made.}
\label{subfig:tweets_by_tag}
  \end{subfigure}

\caption{}
\end{figure}

Based on this analysis, we have to decide which tweets are considered humorous. Let us define the tweets considered humorous as \emph{positives} and the ones considered as non-humorous as \emph{negatives}. The decision consisted in marking as positives those tweets whose ratio of humorous annotations is greater than or equal to 0.6 and as negatives those lower than or equal to 0.3. The rest are considered as \emph{doubtful}. The criterion of giving a 0.1 handicap to the positives was thereby performed, as they are obtained from humorous accounts. This may be seen as if the source is giving its opinion too. Additionally, those tweets with no annotations fall into the category of doubtful. \Cref{subfig:decision} illustrates the proportions of each category. To sum up, 5,952 tweets are considered positive. The rest of the tweets obtained from humorous accounts are not taken into account, even though the negatives can also be used. The corpus composition is shown in \cref{subfig:tweets_by_tag}.

\subsection{Classifier}
\label{subsec:classifier}

Firstly, we split data into 80\% for training and 20\% for later evaluation. Similarly to the works mentioned in this document, we built a humor classifier but for the Spanish language. Such works used Support Vector Machine (SVM) and a Multinomial version of Naïve Bayes (MNB). However, more machine learning techniques are tried here: Decision Trees (DT), k Nearest Neighbors (kNN) and a Gaussian version of Naïve Bayes (GNB). Tweets are tokenized using Freeling~\cite{padro12}. Also, a higher quantity of features was implemented, which is described below.\footnote{The codebase for the classifier and the corpus built can be found in \url{https://github.com/pln-fing-udelar/pghumor}.}

\begin{description}
  \item[Adult slang:] Here we count the relative number of tokens in the tweets which appeared in a previously built dictionary about adult slang. This dictionary contains 132 words, and it was built using bootstrapping, in a similar manner to~\textcite{mihalcea2005bootstrapping}, with a seed of 21 words. Dictionary-lookup features are computed with this formula (where the multiset intersection is used):

  \[
    featureValue(tweet) = \frac{|tweet \cap dictionary|}{\sqrt{|tweet|}}
  \]

  \item[Animal presence:] In this case we compare against a handcrafted dictionary about animals. This dictionary contains 103 names, including typical typographic misspellings and grammatical mistakes.

  \item[Antonyms:] Given a tweet, this feature counts the relative number of pairs of antonyms existing in it. WordNet~\cite{wordnet} antonymy relationship and Spanish language enrichment provided by the Multilingual Central Repository~\cite{mcr} are used for this. This feature was discarded since after performing Recursive Feature Elimination~\cite{guyon2002gene} (RFE) we found out the classification worsened.

  \item[Dialog:] This feature only establishes if a tweet is a dialog.

  \item[Exclamations:] The relative number of exclamation marks are counted.

  \item[First and Second person:] These two features try to capture verbs conjugated in the first and second persons and nouns and adjectives which agree with such conjugations (in Spanish, nouns and adjectives express gender and number at the end of the word).

  \item[Hashtags:] The amount of hashtags in the tweet is counted. It is suspected that the higher this amount is, the more informal the tweet is. Thus, it is more likely to be humorous.

  \item[Keywords:] An intuitively handmade dictionary of 43 common words found in jokes was built for this, and it was used for checking purposes.

  \item[Links:] This feature counts the number of links contained in a tweet.

  \item[Negation:] Here we count the relative quantity of times the word ``no'' appears in the tweet. It was removed after running RFE\@.

  \item[Non-Spanish words:] The relative number of words containing non-Spanish words is counted. It was discarded after running RFE\@.

  \item[Out of vocabulary:] The idea behind this is to keep record of the relative count of words not found in dictionaries. These are four features based on the combination of the dictionaries used: Freeling, Freeling-Google~\footnote{\url{https://www.google.com}}, Freeling-Wiktionary~\footnote{\url{https://www.wiktionary.org}} and Wiktionary.

  \item[Questions-answers:] One interesting attribute for tweets is to count how many questions and answers are present, one after another.

  \item[Topic distance:] The idea is to check if a tweet is somewhat near to a joke category in \emph{Chistes.com}, or whether it is closer to a Wikipedia's sentence, from Wikicorpus~\cite{reese10}. This is carried out using a Multinomial Naïve Bayes classifier together with the Bag of Words technique.

  \item[Uppercase words:] The relative amount of words completely in uppercase is counted.
\end{description}

\section{Experimental Evaluation}
\label{sec:evaluation}

Provided that our work is the only one using this corpus, and even the only one with the goal of classifying humor in Spanish, we cannot directly compare it with any other work. Hence, we developed two baselines to compare it with, aiming them to be simple ideas which could be crafted to face this task. The first one (BL1) is a Multinomial Naïve Bayes classifier combined with Bag of Words similarly to the Topic Distance feature. The second one (BL2) is a classifier which predicts all tweets with the most likely outcome, \emph{non-humorous}, having a frequency of almost 83\%.

A comparison using mainly the $F_1$ score is intended. We want to pay attention to the positives (the humorous) but also granting the same degree of importance to false positives and false negatives. Nonetheless, we take advantage of the runs in order to also pay attention to other measurements. The results are shown in \cref{tab:results}.

\begin{table}
  \centering
  \caption{Results obtained with the different techniques over the test set. NPV, TNR and Neg.\ F1 refer to Precision, Recall and $F_1$ score, respectively, when reversing the roles positive-negative.}
  \begin{tabular}{c r r r r r r r}
    \toprule
    & \multicolumn{1}{c}{Precision} & \multicolumn{1}{c}{Recall} & \multicolumn{1}{c}{$F_1$} & \multicolumn{1}{c}{NPV} & \multicolumn{1}{c}{TNR} & \multicolumn{1}{c}{Neg.\ $F_1$} & \multicolumn{1}{c}{Accuracy} \\
    \midrule
    BL1 & 0.617 & 0.846 & 0.714 & 0.966 & 0.892 & 0.714 & 0.885 \\
    BL2 & N/A & 0.000 & N/A & 0.830 & 1.000 & 0.907 & 0.830 \\
    \midrule
    SVM & 0.836 & 0.689 & \textbf{0.755} & 0.938 & 0.972 & \textbf{0.955} & \textbf{0.925} \\
    DT & 0.665 & 0.675 & 0.670 & 0.933 & 0.930 & 0.932 & 0.889 \\
    GNB & 0.575 & \textbf{0.782} & 0.663 & \textbf{0.952} & 0.882 & 0.915 & 0.865 \\
    MNB & \textbf{0.848} & 0.600 & 0.703 & 0.923 & \textbf{0.978} & 0.950 & 0.914 \\
    kNN & 0.813 & 0.663 & 0.730 & 0.934 & 0.969 & 0.951 & 0.917 \\
    \bottomrule
  \end{tabular}
\label{tab:results}
\end{table}

The best results are obtained with SVM, even in terms of accuracy. Also, kNN shows satisfactory output. These two approaches outperform the baselines, with the former clearly surpassing the latter. Meanwhile, GNB and DT have poor precision, although GNB certainly does a better job among these two and has the best recall. The confusion matrix for SVM is shown in \cref{tab:matrix}.

\begin{table}
  \centering
  \caption{Confusion matrix for SVM classifier with respect to the test set}
  \begin{tabular}{c r r}
    \toprule
     & Positive & Negative \\
    \midrule
    Positive & 842 & 381 \\
    \midrule
    Negative & 165 & 5805 \\
    \bottomrule
  \end{tabular}
\label{tab:matrix}
\end{table}

\section{Conclusions}
\label{sec:conclusions}

A crowdsourced corpus has been assembled, which serves the purpose of this work and could be useful for future research. It contains over 30,000 annotations for 16,488 tweets, coming from humorous accounts, and it also counts with 22,875 sourced from non-humorous accounts. Uses of such corpus include analyzing its data, as well as performing tasks similar to the work described herein.

We have built a classifier which outperforms the baselines outlined. Support Vector Machine proved to be the best technique. It has a precision of 83.6\%, a recall of 68.9\%, a $F_1$ score of 75.5\% and an accuracy of 92.5\%. Nevertheless, it must be highlighted that the corpus built does not depict a great variety of humor. Hence, some features perform well in this work but might not perform so well in another context.

As a future work, more complex features could be crafted, such as trying to detect wordplay and puns, ambiguity, perplexity against some language model, inter alia. Other Machine Learning techniques could also be tried. It would be interesting if we take advantage of the star ranking people provided; maybe this can also suggest how funny a joke is. As a harder task, humor generation could be tackled. Finally, it could be studied how the influence of humor varies between different social contexts, depending on gender, age, interest areas, mood, etc.

\printbibliography 

\end{document}